# Outlier Detection in Plantar Pressure: Human-Centered Comparison of Statistical Parametric Mapping and Explainable Machine Learning

Carlo Dindorf[1,*], Jonas Dully[1], Steven Simon[1], Dennis Perchthaler[1], Stephan Becker[1], Hannah Ehmann[2], Kjell Heitmann[2], Bernd Stetter[3], Christian Diers[2], Michael Fröhlich[1]

[1] Department of Sports Science, University of Kaiserslautern-Landau (RPTU), Kaiserslautern, Germany

[2] DIERS International GmbH, Wiesbaden, Germany

[3] Institute of Sports and Sports Science, Karlsruhe Institute of Technology, Karlsruhe, Germany

* Correspondence:
Carlo Dindorf
carlo.dindorf@rptu.de

**Abstract:**

Plantar pressure mapping is essential in clinical diagnostics and sports science, yet large heterogeneous datasets often contain outliers from technical errors or procedural inconsistencies. Statistical Parametric Mapping (SPM) provides interpretable analyses but is sensitive to alignment and its capacity for robust outlier detection remains unclear. This study compares an SPM approach with an explainable machine learning (ML) approach to establish transparent quality-control pipelines for plantar pressure datasets. Data from multiple centers were annotated by expert consensus and enriched with synthetic anomalies resulting in 798 valid samples and 2000 outliers. We evaluated (i) a non-parametric, registration-dependent SPM approach and (ii) a convolutional neural network (CNN), explained using SHapley Additive exPlanations (SHAP). Performance was assessed via nested cross-validation; explanation quality via a semantic differential survey with domain experts. The ML model reached high accuracy and outperformed SPM, which misclassified clinically meaningful variations and missed true outliers (Matthews Correlation Coefficient: ML = 0.96 ± 0.01; SPM = 0.78 ± 0.02). Experts perceived both SPM and SHAP explanations as clear, useful, and trustworthy, though SPM was assessed less complex. These findings highlight the complementary potential of SPM and explainable ML as approaches for automated outlier detection in plantar pressure data, and underscore the importance of explainability in translating complex model outputs into interpretable insights that can effectively inform decision-making.

# 1. Introduction

Plantar pressure mapping—capturing how vertical forces distribute across the foot during static (e.g., standing) or dynamic (e.g., walking, running) activities—has become indispensable in clinical diagnostics, sports science, and rehabilitation [1–3]. By revealing biomechanical irregularities in pressure profiles, the diagnosis, monitoring, and screening of conditions such as diabetic neuropathy [4], Parkinson's disease [5], and various musculoskeletal disorders [6–8], is supported. It is also an established approach to measure the biomechanical impact of medical aids, such as (knee) ankle-foot orthotics or insoles, to ensure positive clinical outcomes [9] and in the recent past to adapt orthotics to plantar pressure profiles [9,10]. Yet the accuracy of any downstream analysis hinges on data quality—and in practice, pressure datasets are often contaminated by outliers, or anomalies, which are data points that deviate from the expected pattern [11].

Variations in participant instruction, protocol adherence, or the use of different systems across multiple centers are causes for inconsistencies and increase the likelihood of outliers in plantar pressure data. Instructor- or protocol-related issues may include incorrect guidance, inadequate monitoring, or failure to correct participant behavior (e.g., wearing unintended footwear or moving unexpectedly during trials). Technical factors include sensor malfunctions, algorithmic misalignment (e.g., on treadmills), premature truncation of pressure curves, environmental noise, or errors in automated foot identification and segmentation. If left unaddressed, such outliers can distort automated segmentation algorithms [12,13] and reduce the accuracy of classification pipelines [14,15], potentially leading to erroneous or lower-quality biomechanical insights. This issue is particularly critical in multicenter data collection, which is often required to overcome data sparsity, especially when developing automated machine learning (ML) pipelines [16]. In such settings, datasets can become large and heterogeneous, making manual data verification economically and logistically impractical.

One established approach for analyzing plantar pressure data is Statistical Parametric Mapping (SPM) [17]. Originally developed for the analysis of 3D neuroimaging data [18], SPM enables the statistical comparison of continuous spatial data by performing voxel- or point-wise variance analyses across pre-defined regions. In the context of plantar pressure, this enables researchers to identify areas of statistically significant deviation across the plantar surface rather than relying solely on summary metrics such as peak pressure alone [17]. Frameworks such as Personalized Analysis of Plantar Pressure Images (PAPPI) [19] leverage SPM by first constructing a normative model that accounts for individual demographic factors (e.g., age, weight, or foot size) and then applying SPM to highlight deviations from this normative reference. While PAPPI showcases the interpretability and individualized assessment strengths of SPM, it was not explicitly designed to isolate technical errors or procedural inconsistencies; instead, it flags any departure from the norm, independently whether pathological or spurious. This limitation underscores a critical requirement of SPM analyses: spatial alignment. To ensure that each pixel corresponds to the same anatomical region across subjects, plantar pressure data must be normalized for rotation, scale, and anatomical landmarks [1]. Without such alignment, statistically significant deviations may reflect misregistration rather than true biomechanical differences—for example, identical pixel locations could span both heel and midfoot areas in different participants, leading to misleading inferences.

ML methods offer a promising alternative, as they can potentially learn alignment invariances directly from the data, reducing or even eliminating the need for labor-intensive preprocessing steps [20]. In medical imaging, deep learning approaches have demonstrated high-precision anomaly detection [21,22]. However, it remains largely unexplored whether these methods can achieve similar success in plantar pressure data, particularly when trained on labeled examples of diverse outlier types.

Importantly, under Article 22 of the General Data Protection Regulation (GDPR), individuals subjected to automated decision-making have the right to obtain meaningful information about the logic underlying such decisions [23]. This requirement presents a significant challenge for ML models, which often operate as "black boxes" that do not readily provide interpretable explanations. To address this challenge, Explainable Artificial Intelligence (XAI) techniques have gained increasing importance, enabling researchers and practitioners to probe the internal decision-making processes of complex ML models [24]. Beyond facilitating model debugging, XAI methods support risk assessment, bias detection, regulatory compliance, and the development of end-user trust and acceptance [25].

Crucially, the value of an explanation extends beyond quantitative metrics (e.g., fidelity or completeness) to encompass human-centered attributes, including understandability, reliability, and the potential to inform subsequent actions [26,27]. Without systematic, human-in-the-loop evaluation, XAI outputs risk relegation to academic curiosities rather than serving as practical decision-support tools in clinical settings.

Despite recent advances in the field of explainable outlier detection [28], the integration of supervised outlier classification with XAI—along with human-centered assessment of explanatory outputs—has, to the best of the authors' knowledge, not yet been explored in plantar pressure data. This constitutes a critical research gap, given the growing need for automated yet interpretable quality-control pipelines in clinical and sports analysis [29]. To address this gap, this study directly compares the more established SPM approach against a novel explainable ML approach. This investigation is structured around two core questions:

1) How do these approaches compare in terms of detection accuracy?
2) In an exploratory follow-up, how do human evaluators perceive and trust the explanations generated by each approach?

By addressing these questions, this study aims to inform strategies for refining data-cleaning protocols and guiding the development of real-time monitoring systems that can alert technicians to potential acquisition errors (e.g., prompting trial repetition or verifying foot-side annotations). Ultimately, these advancements are intended to enhance data quality and reliability in both research and practical diagnostic settings, specifically within the context of plantar pressure analysis.

# 2. Methods

## 2.1. Workflow overview

An overview of the workflow is presented in Figure 1 and described in more detail below.

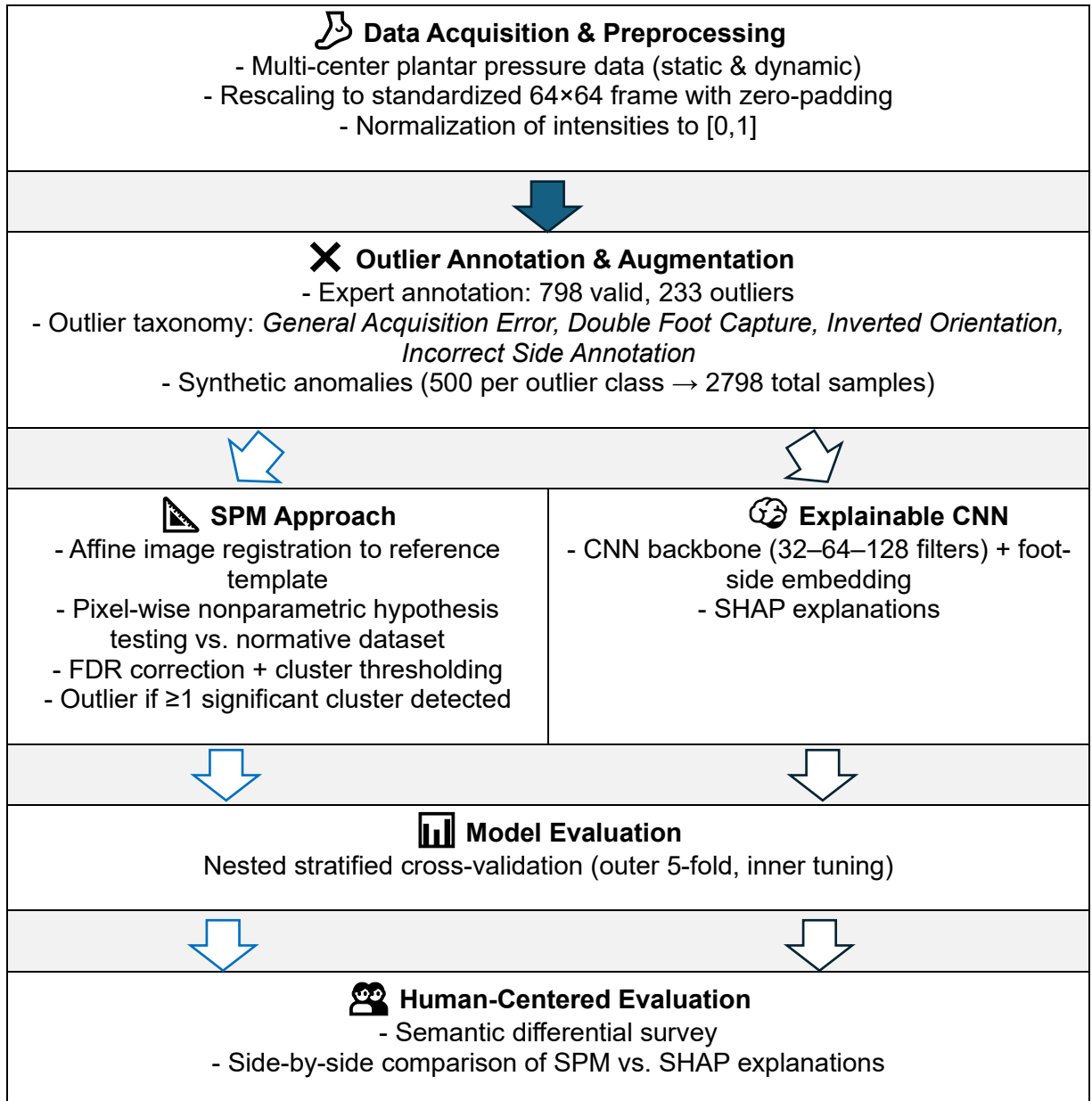

**Figure 1.** Overview of the study workflow, illustrating the comparison between Statistical Parametric Mapping (SPM) and a machine learning approach based on a Convolutional Neural Network (CNN), with model interpretation provided through SHapley Additive exPlanations (SHAP).

## 2.2. Participants and Data Acquisition

Participation in the study was restricted to individuals of legal age. All participants received detailed information about the study protocol and relevant data protection guidelines before giving their written informed consent. The study was conducted in accordance with the ethical standards set forth in the Declaration of Helsinki and received approval from the Ethics Committee of the University of Kaiserslautern-Landau (approval number: 55). Data were collected across several centers using two types of pressure measurement systems: resistive pressure sensor plates (RSscan Lab Ltd., Ipswich, England) and capacitive pressure sensor plates (Zebris Medical GmbH, Isny, Germany). Measurements were taken from both the left and right feet under static (53%) and dynamic (47%) conditions. In the static trials, participants stood barefoot in an upright position on the platform for 10 seconds, with data captured at 50 Hz. Following a 60-second habituation phase on a treadmill, dynamic trials consisted of three overground walking passes at each participant's self-selected pace, recorded at 100 Hz. For these dynamic measurements, stride-level plantar pressure profiles and peak pressure values were extracted, in line with common reporting practices [3]. Duplicate datasets were identified and excluded before analysis.

Only data essential to the development and evaluation of the computational model were collected, while anthropometric and other descriptive variables were deliberately omitted. This choice reflected the principle of data minimization (GDPR Art. 5(1)(c) [23]), requiring personal data to be "adequate, relevant, and limited to what is necessary" for the stated purpose.

Because the two pressure systems differed in spatial resolution and sensor geometry, preprocessing steps were applied to harmonize the datasets. First, all pressure distributions were rescaled to adjust for non-uniform sensor spacing. The resulting data were proportionally resized and embedded into a standardized 64 × 64 pixel grid, maintaining aspect ratios and avoiding distortion by applying zero-padding around the patterns. Finally, pressure intensities were normalized to the range [0,1] to reduce inter-subject variability and eliminate weight-dependent effects.

## 2.3. Outlier Annotation and Dataset Augmentation

Outlier categories were defined by domain experts through systematic review of the dataset, considering both naturally occurring artifacts and recurrent sources of error in the initial dataset. The resulting taxonomy is presented in Table 1. The categorization was guided not only by technical accuracy but also by the practical relevance of providing automated feedback to end users regarding data integrity. Specifically, recordings classified as *General Acquisition Errors* were consolidated into a single category, because such trials are irreparably flawed (e.g., incomplete foot contact, trials performed with footwear, or corrupted sensor output) and require re-acquisition rather than post hoc correction. In contrast, the remaining categories capture systematic but potentially correctable errors—such as mislabeling of foot laterality or inverted foot orientation—that could be addressed through post-processing.

If a sample was identified as having both an *Incorrect Side Annotation* label and another outlier category, the *Incorrect Side Annotation* was considered the lowest priority and was superseded by the more critical label. This method was implemented to ensure that each sample was assigned to the most practically relevant outlier category, which aids in subsequent actions, such as deciding whether re-recording the data is required.

**Table 1.** Overview of the data categories included in the dataset.

| Label | Example | | Description |
|---|---|---|---|
| **Valid Sample**<br><br>**Label 0** | 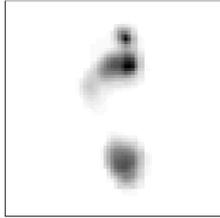<br>Left Side | 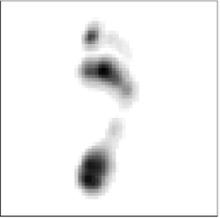<br>Right Side | **Correct plantar pressure recordings without measurement or detection errors**<br><br>Both feet are properly identified, and side labels (left vs. right) are accurate. Importantly, atypical or pathological plantar pressure patterns (e.g., due to gait abnormalities) are not classified as outliers if acquisition quality is intact. |
| **General Acquisition Error**<br><br>**Label 1** | 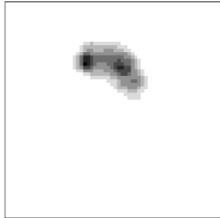<br>Right Side | 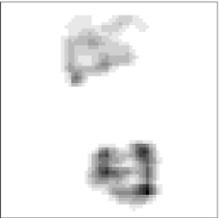<br>Left Side | **Recordings with severe acquisition artifacts**<br><br>These include incomplete foot contact (e.g., only forefoot or heel captured), trials performed with footwear instead of barefoot, or corrupted sensor output (e.g., motion blur, hardware malfunction). |
| **Double Foot Capture**<br><br>**Label 2** | 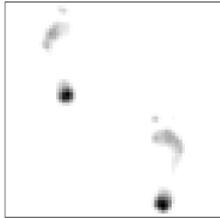<br>Right Side | 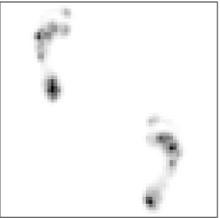<br>Right Side | **Failure to separate left and right plantar pressure distribution**<br><br>Trials in which both feet are in a single frame due to failed automated segmentation of individual footprints. |
| **Inverted Orientation**<br><br>**Label 3** | 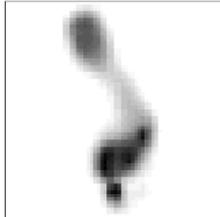<br>Left Side | 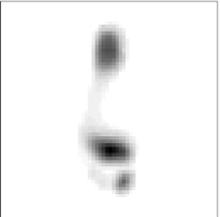<br>Right Side | **Non-standard orientation of plantar pressure maps**<br><br>Plantar pressure maps with non-standard orientation, where the forefoot is not aligned upwards, e.g. due to incorrect foot placement on the pressure plate or erroneous test leader instructions. |
| **Incorrect Side Annotation**<br><br>**Label 4** | 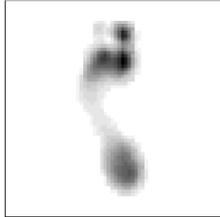<br>Right Side | 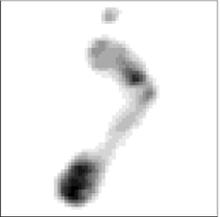<br>Left Side | **Incorrect side labeling**<br><br>Plantar pressure distribution is valid, and acquisition quality is intact, but metadata incorrectly labels the side of the foot, e.g., right foot recorded but annotated as left. |

Three domain experts focused on identifying and labeling anomalous (outliers) and valid (inliers) samples. The annotation process was carried out collaboratively. Each sample was independently reviewed by at least one expert, after which labels were cross-verified to ensure consensus across raters. This process yielded a curated dataset of 1,031 samples from 703 subjects, consisting of 798 valid samples and 233 outlier samples (*General Acquisition Error*:

124; *Double Foot Capture*: 29; *Inverted Orientation*: 38; *Incorrect Side Annotation*: 42), with approximately equal representation of left and right feet (~50% each). To improve model robustness and allow for systematic evaluation, the dataset was further augmented with synthetically generated outliers, ensuring that each outlier category contained 500 samples. This methodological decision is supported by prior works in other fields, which have shown performance gains when balancing outlier classes using synthetic data [30]. Four types of artificial outliers were created:

- ***General Acquisition Error* (Label 1):** Simulated by spatially cropping plantar pressure maps to remove either the forefoot or heel regions, thereby mimicking incomplete foot contact. Original laterality annotations were retained.

- ***Double Foot Capture* (Label 2):** Constructed by merging left and right foot pressure maps from different individuals. Domain experts guided alignment and positioning to preserve anatomical plausibility. A random laterality label was assigned.

- ***Inverted Orientation* (Label 3):** Generated by rotating inlier pressure maps by 180°, simulating upside-down foot placement. Original laterality annotations were retained.

- ***Incorrect Side Annotation* (Label 4):** Introduced by randomly inverting left/right labels of valid inlier samples.

The fidelity of these synthetic outliers was evaluated by the three experts, who confirmed that the artificially generated samples closely reproduced the biomechanical and acquisition-related characteristics of genuine outliers within each category. After augmentation, the final dataset comprised 2798 samples in total, reflecting 798 inliers and 2000 outliers.

## 2.4. Statistical Parametric Mapping (SPM) Approach

### 2.4.1. Plantar Pressure Registration

A critical prerequisite for the SPM approach is that all images be precisely aligned to ensure that each pixel corresponds to the same anatomical foot region across all subjects, thereby preventing misalignment from introducing misleading statistical inferences [17]. To address this, a plantar pressure registration pipeline was implemented using a similarity-based optimization method, which has previously been shown to be effective for plantar pressure alignment [31].

To ensure spatial consistency, each raw input was registered to a single, pre-defined prototypical reference pressure distribution, generated separately for the left and right foot. This reference acts as a template for alignment. The registration process employed an affine transformation, which corrects for variations in rotation, translation, and scaling [32]. The optimal transformation parameters (angle, shift, and zoom) were found by minimizing a mean squared error (MSE) loss function between the transformed input and the corresponding prototypical reference, following established image registration procedures [33]. The optimization was performed using the L-BFGS-B algorithm [34], ensuring that the transformations were constrained within realistic bounds. Figure 2 shows a visual example of the registration results.

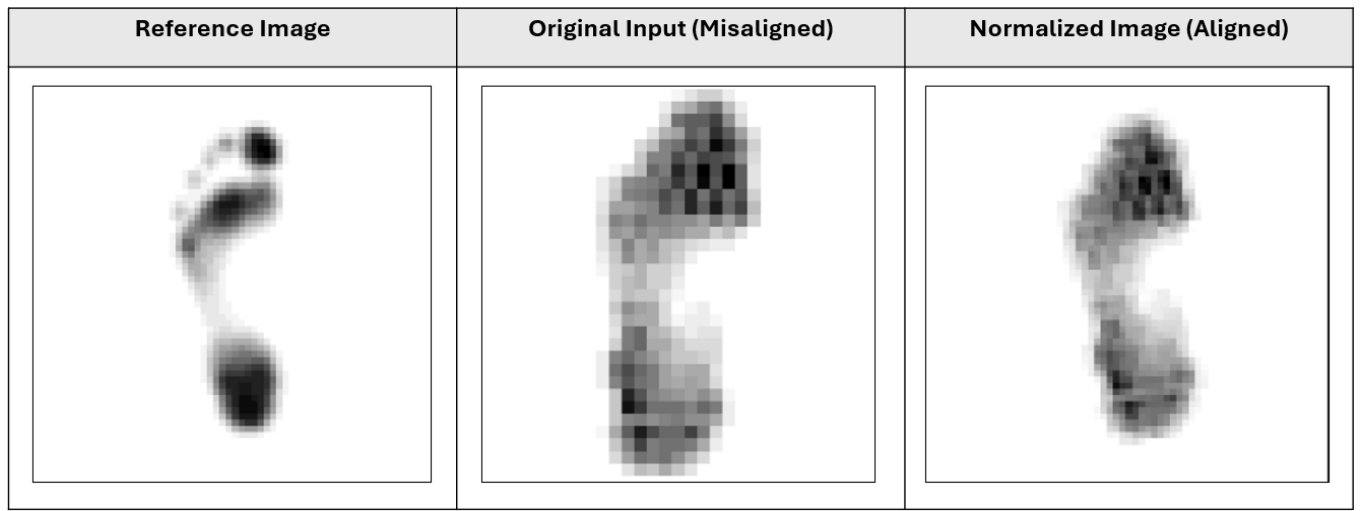

**Figure 2.** Exemplary plantar pressure registration results.

### 2.4.2. Statistical Analysis

We implemented a non-parametric SPM approach tailored for plantar pressure data, using cluster-based permutation testing to identify deviant (outlier) pressure maps. This approach extends the widely used methodology of non-parametric SPM [35,36] to the specific task of technical error or procedural inconsistency detection in plantar pressure data. The analysis is grounded in a normative modeling framework, where each test sample is statistically compared against a reference distribution derived from a cohort of non-outlier (healthy) plantar pressure maps. Because left and right feet differ in anatomy and loading patterns, analyses were conducted independently for each laterality, preventing anatomical misalignment even after spatial normalization.

Our methodology proceeds in two main steps:

1. **Pixel-wise non-parametric testing:** For each pixel in a test sample, we compute a two-tailed empirical p-value by comparing its pressure intensity against the empirical distribution at the same pixel across the normative reference cohort. This is achieved via a rank-based permutation approach [36], which is robust to non-Gaussian distributions and does not rely on parametric assumptions. The resulting p-map represents the probability of observing such an extreme pressure value under the normative model.

2. **Cluster-based multiple comparison correction:** To control the large number of simultaneous pixelwise tests, we employed a cluster-based permutation procedure [35]. First, clusters of contiguous suprathreshold pixels were identified using an uncorrected cluster-forming threshold ($\alpha\_forming$). Next, a null distribution of maximum cluster sizes was constructed by repeatedly permuting the normative data (n = 1,000 permutations) and recomputing the p-map, thereby quantifying the cluster sizes expected under the null hypothesis. Finally, only clusters whose size exceeded the (1 − $\alpha\_FWE$) percentile of the null distribution (here, $\alpha\_FWE = 0.05$) were retained as significant. This controls the family-wise error rate at the cluster level, ensuring that the probability of one or more false positive clusters is bounded by $\alpha\_FWE$.

To increase robustness, we additionally required clusters to meet a minimum cluster size (min_cluster) criterion, which served as an initial filter before running the full permutation-based correction. The two key parameters of this—$\alpha\_forming$ (cluster-forming threshold) and min_cluster (minimum cluster size)—were tuned in a nested cross-validation scheme (see section 2.6). Specifically, a randomized search was performed in the inner loop of the cross-validation, where α_forming was sampled uniformly from the range 0.01–0.05, and min_cluster from the range 0–30 pixels. Candidate parameter sets were evaluated on the validation folds using the F1-score (binary inlier vs. outlier classification), and the best-performing combination was carried forward to the outer cross-validation loop for final evaluation.

## 2.5. Machine Learning Approach

A Convolutional Neural Network (CNN) classifier designed to process the plantar pressure data along with additional categorical metadata (foot lateral label) was implemented [37,38]. Parameter search was performed manually by evaluating the model's performance on the validation sets. Starting with a more complex architecture, the model's complexity was gradually reduced until further reductions led to worse validation performance. The resulting CNN backbone consists of three sequential convolutional blocks with increasing filter sizes (32, 64, 128). Each block is composed of two convolutional layers with a 3x3 kernel, followed by batch normalization, a ReLU activation function, and a 2x2 max-pooling layer. Dropout layers (Dropout2D) were included after each max-pooling operation to mitigate overfitting.

The flattened output of the convolutional layers is concatenated with a learned embedding vector from a categorical embedding layer. This layer converts the integer-encoded side labels into an 8-dimensional vector representation. The combined vector is then passed to a classifier head comprising two fully connected layers with batch normalization and dropout. The final layer outputs the class probabilities.

The plantar pressure samples were normalized using the mean and standard deviation calculated exclusively from the training data of each cross-validation fold (see section 2.6) to prevent data leakage. The categorical variable was encoded as an integer and passed directly to the embedding layer. The data were structured using a custom dataset class, with class-balanced sampling to address label imbalance during training.

The model was trained using a weighted cross-entropy loss, where class weights were inversely proportional to their frequency in the training set. The Adam optimizer with a learning rate of 0.001 was used for weight updates. Training was managed with an early stopping mechanism, which halted the process if the validation loss did not improve for ten consecutive epochs, and the model with the lowest validation loss was retained.

To understand the key plantar presser regions that contributed to the CNN's predictions, we employed the XAI method SHapley Additive exPlanations (SHAP) [39]. This method explains the output of a ML model as a sum of contributions from each input feature, providing local interpretability. We used a Deep SHAP explainer, which is tailored for deep learning models. The explainer was trained on a background dataset of a random subset of 100 plantar pressure samples and their corresponding metadata from the training set. For each test sample, the SHAP explainer calculated the contribution of each sample pixel to the final prediction.

## 2.6. Evaluation and Calculations

To ensure an unbiased assessment of the two approaches, we employed nested stratified cross-validation. The nested structure is critical for preventing data leakage and ensuring that the final performance metrics accurately reflect the models' generalization ability on unseen data [40]. The same data partitions were used for both approaches, enabling a direct and fair comparison of their performance. An outer 5-fold stratified cross-validation was used to partition the dataset into a training/validation set and a held-out test set. Within each outer training/validation fold, an inner stratified shuffle split (80/20 ratio) was applied to create a dedicated training set and a validation set for hyperparameter tuning. For both the SPM and ML approaches, the grouped data structure was respected, ensuring that all data from a single subject (e.g., both left and right foot) were confined to a single partition. This is important to mitigate the risk of artificially inflated performance due to anatomical or measurement similarities within a subject.

The best-performing model configuration—identified by its performance on the validation set during the inner loop—was then evaluated on the completely independent, held-out test set. This process was repeated for each fold of the outer loop. The final model's performance was assessed using the Matthews Correlation Coefficient (MCC) and F1-score, which are robust metrics for evaluating models in the presence of class imbalance [41,42]. To ensure a fair comparison, the multi-class predictions from the ML approach were additionally post-processed into a binary classification output, analogous to the SPM-inspired predictions. In addition, confusion matrices were generated to visualize class-wise prediction accuracy and to identify potential sources of bias. All modeling, training, and evaluation procedures were implemented in Python using PyTorch [43], scikit-learn [44], and SciPy [45], while visualizations were generated with Matplotlib [46] and Seaborn [47].

## 2.7. Human-Centered Results Evaluation

### 2.7.1. Visual Representation

To provide a comprehensive understanding of the models' decision-making processes, a side-by-side visualization of the outputs from the SPM and the ML approach is provided. For each sample analyzed, we generated a three-panel figure. The leftmost panel displays the original grayscale plantar pressure. The central panel presents the output of the non-parametric SPM approach. Here, the original pressure is shown with a green contour line overlaying regions that were identified as statistically significant outliers according to the approach. This highlights the specific foot regions where pressure values deviate substantially from the normative, valid plantar pressure dataset.

The rightmost panel presents the explanation of the CNN model's predictions using SHAP values. It overlays a heatmap on the original plantar pressure, highlighting pixels that positively or negatively contributed to the model's output. This provides a visual representation of the most influential plantar pressure regions underlying the predicted classification. To enhance clarity, SHAP values below 20% of the maximum absolute value were omitted, and the resulting map was smoothed using a bilateral filter.

### 2.7.2. Semantic Differential Survey

For this exploratory part of the study, we recruited 16 participants (9 male, 7 female) with expertise in biomedical data analysis and plantar pressure assessment. All participants held at least a university degree in sports science, biomechanics, or a health-related field, and reported extensive prior experience with plantar pressure data. The participants were provided with written, standardized explanations on how to interpret the explanations provided. The presentation to participants and data collection were carried out using the digital survey platform LimeSurvey (LimeSurvey GmbH, Hamburg, Germany). The estimated duration for completing the entire survey was approximately 25 minutes.

To explore how end-users perceive the quality of explanations generated we employed a semantic differential survey. A semantic differential is a well-established psychometric technique in which respondents rate a concept along bipolar adjective scales, thereby yielding quantitative measures of subjective impressions [48].

Drawing on prior work in human-centered XAI and established criteria for evaluating explanation quality, we selected eight key attribute pairs that capture core dimensions relevant to users' understanding and trust in AI explanations [26,27,49]. These pairs were selected to evaluate two key aspects of explanations: their cognitive processing and their perceived utility.

Each of the following adjective pairs was presented on a 7-point Likert scale, with the two extremes representing the poles of the pair:

- **Understandability** (Understandable – Unintelligible)
- **Correctness** (Correct – Incorrect)
- **Trustworthiness** (Trustworthy – Suspicious)
- **Usefulness** (Useful – Useless)
- **Clarity** (Clear – Unclear)
- **Completeness** (Complete – Incomplete)
- **Simplicity** (Simple – Complex)
- **Relevance** (Relevant – Irrelevant)

The left-to-right order of the attributes (e.g., "Correct-Incorrect" vs. "Incorrect-Correct") was randomized for each participant to minimize bias. Each approach was evaluated separately using the semantic differential scale. The presented explanations were limited to correctly classified samples in both approaches. This decision was made to assess the comparative quality of explanations for accurate model predictions, rather than focusing on the XAI approaches' ability to elucidate misclassifications.

To manage the workload for each participant, we selected a random subset of ten generated explanations to be evaluated by the participants. Each participant was presented with the *same* set of model predictions along with their corresponding explanations from both approaches, as described in Section 2.7.1. This side-by-side presentation allowed for a direct evaluation of the relative strengths and weaknesses of each approach's explanations on the same task. To ensure a fair comparison, we removed the additional level of detail provided by the ML model, which not only indicated whether a sample was predicted as an outlier but also specified the type of outlier. This adjustment was made to avoid bias resulting from differences in the amount of information conveyed by the labels. Furthermore, we labeled the approaches A and B to prevent any bias that could arise from participants knowing which approach was used.

To assess potential statistical differences between the approaches and the attribute ratings in the semantic differential, the Wilcoxon signed-rank test was applied as a non-parametric paired test, due to the ordinal nature of the semantic differential data and the non-normal distribution. For multiple comparisons, p-values were adjusted using the Bonferroni correction. The significance level was set at $\alpha = 0.05$.

To assess the perceived consistency of the explanations, a question was posed to participants for each sample after they had evaluated both approaches. Using a 5-point Likert scale, we asked participants to rate their agreement with the following statement: "Both approaches A and B highlight similar features and reasoning behind the model's classification." This question was designed to gauge the experts' perception of explanation alignment between the two approaches. Finally, for each sample presented, participants were asked, "Which approach would you personally prefer?" Response options included: Approach A, approach B, a combination of both (as presented in the survey), or neither.

# 3. Results

## 3.1. Classification Results

Results are summarized in Table 2. Overall, the ML approach outperformed the SPM approach. A detailed single-case analysis of the misclassified samples revealed that, for both approaches, valid samples incorrectly identified as outliers (false positives) were predominantly feet exhibiting pathological characteristics (e.g., hallux valgus, hammer toe, claw toe, flatfoot). Further analysis of the false negative samples of the SPM approach indicates that samples from the outlier class 4 (*Incorrect Side Annotation*) were most often misclassified as valid (n=109), followed by classes 1 (*General Acquisition Error*; n=49) and class 3 (*Inverted Orientation*; n=29).

A confusion matrix for the ML approach is shown in Figure 3. The lowest label accuracy was observed for the *General Acquisition Error* class. Interestingly, samples belonging to this class were most frequently misclassified as valid samples. For the *Inverted Orientation* class, the primary source of error was misclassification as *General Acquisition Error*.

**Table 2.** Results for the held-out test sets across all cross-validation folds, comparing the SPM and machine learning (ML) approaches. For comparability, predictions of the multiclass ML approach were reduced to a binary classification of outlier vs. non-outlier. The confusion matrices show actual classes on the rows and predicted classes on the columns, where the top row corresponds to valid samples and the bottom row corresponds to outliers. Correct predictions (true positives and true negatives) are highlighted in grey. Metrics include the minimum (min) and maximum (max) values, as well as the mean ± standard deviation (std) for the Matthews Correlation Coefficient (MCC) and F1-score (F1).

|  | SPM Approach | | ML Approach | |
|---|---|---|---|---|
| **Confusion Matrix** | 754 | 44 | 783 | 15 |
|  | 237 | 1763 | 30 | 1970 |
| **MCC (min; max)** | 0.76; 0.81 | | 0.95; 0.98 | |
| **MCC (mean ± std)** | 0.78 ± 0.02 | | 0.96 ± 0.01 | |
| **F1-score (min; max)** | 0.92; 0.94 | | 0.98; 1.00 | |
| **F1-score (mean ± std)** | 0.93 ± 0.01 | | 0.99 ± 0.00 | |

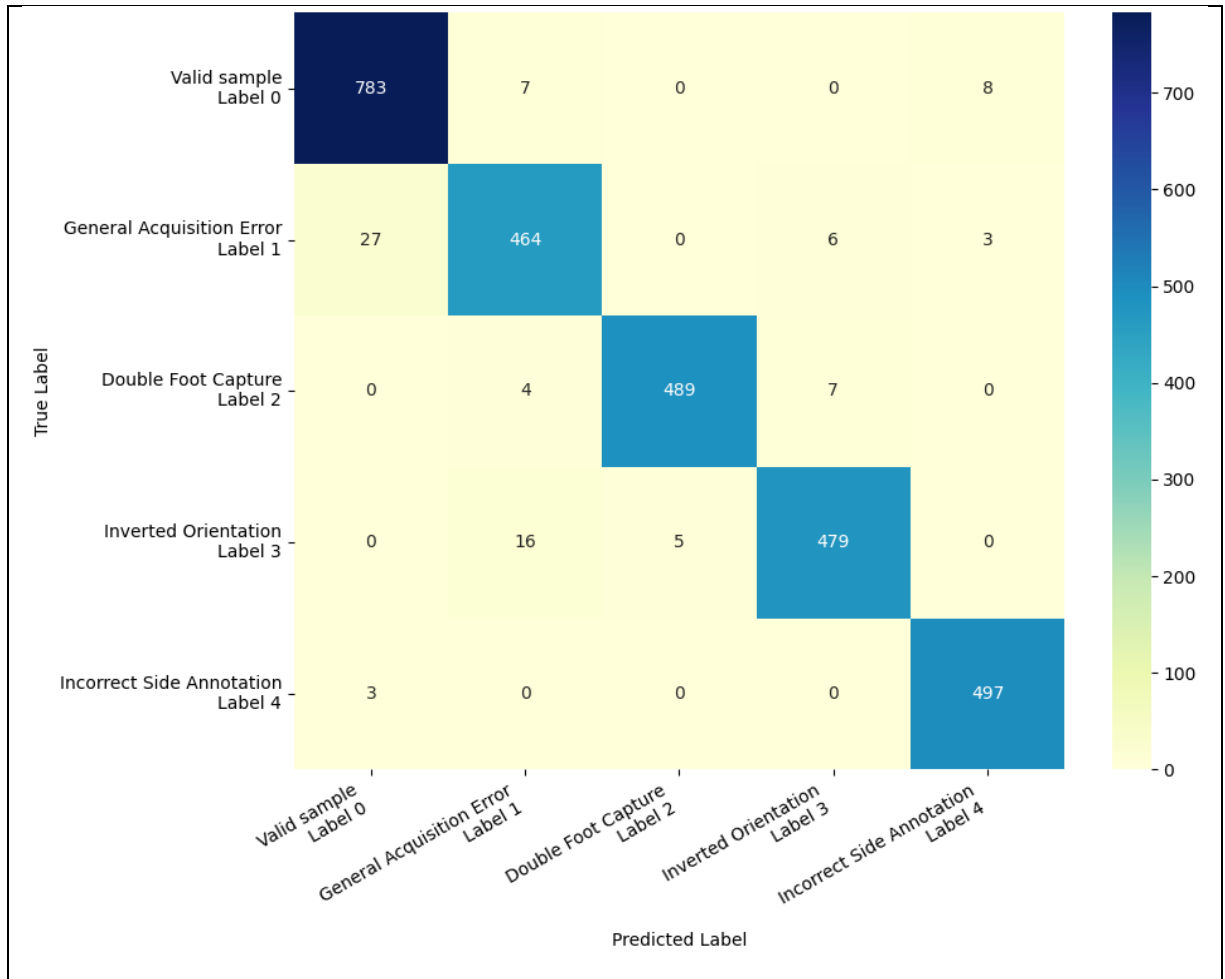

**Figure 3.** Class-specific confusion matrices for the machine learning (ML) approach on the test set.

## 3.2. Semantic Differential Results

Figure 4 presents two exemplary cases of generated explanations using both approaches. While the SPM-approach highlights clusters with statistically significant deviations from the valid dataset, the ML-approach with SHAP explanations highlights areas that contributed to or against the decision.

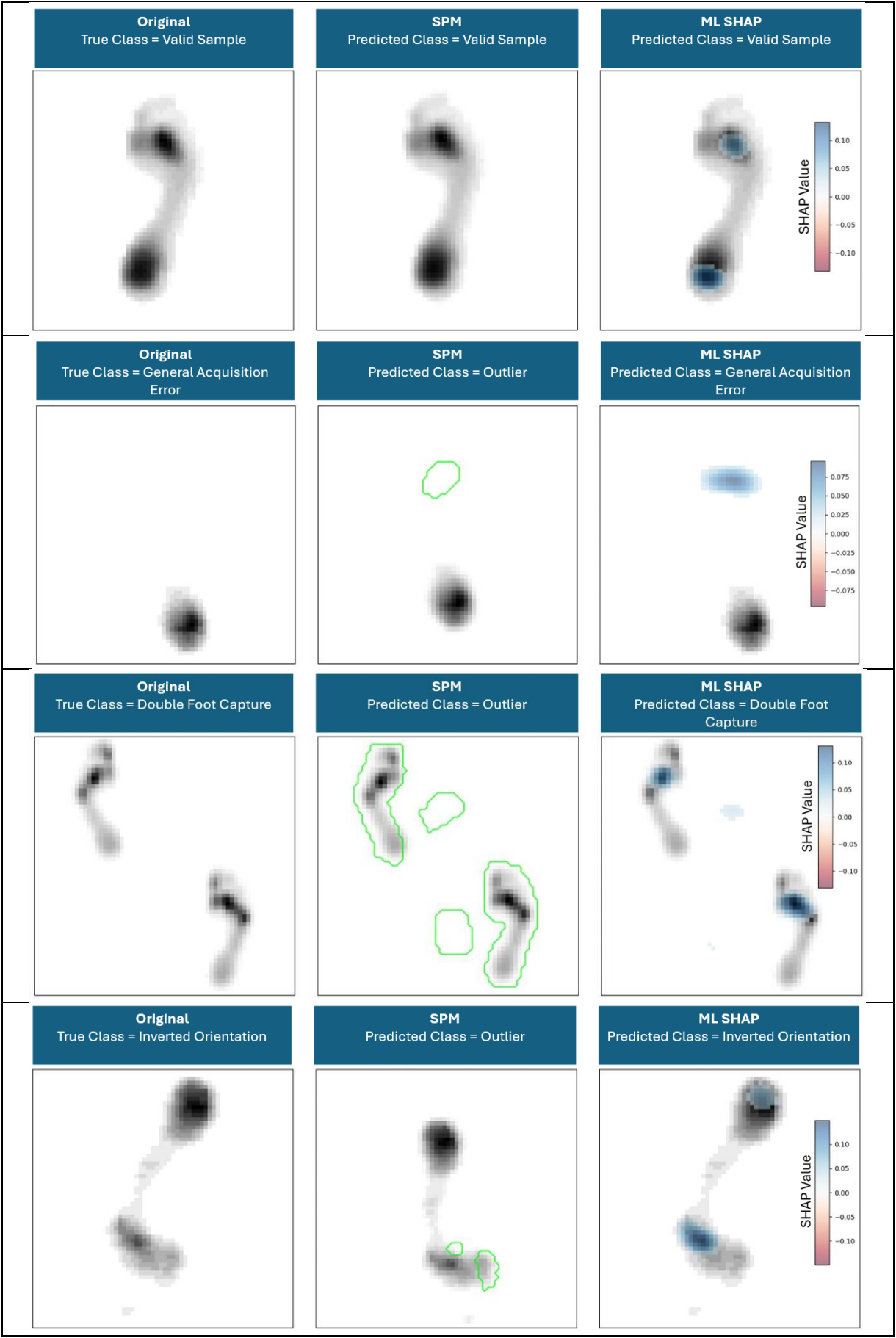

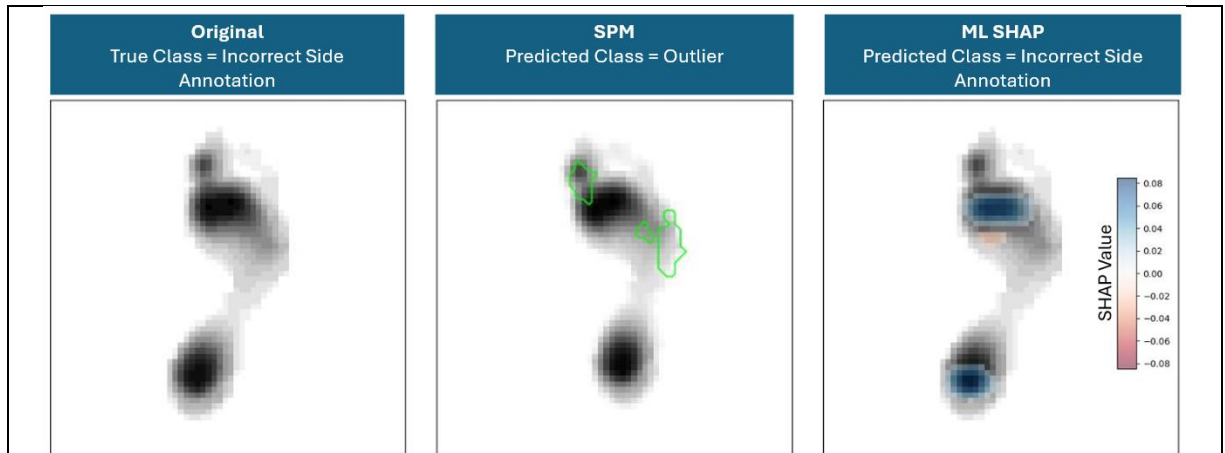

**Figure 4.** Side-by-side visualization comparing outputs from the Statistical Parametric Mapping (SPM) approach and the machine learning (ML) approach for each data category (see Table 1). Each row shows: (left) the original grayscale plantar pressure, (middle) the SPM output with green contours marking regions identified as statistically significant outliers relative to a normative dataset, and (right) the ML explanation using SHAP values, where a heatmap overlays the original pressure distribution to highlight plantar pressure regions with the strongest positive or negative contributions to the model's prediction. Regions colored blue represent pixels that positively contribute to the model's prediction for the classified label, while regions colored red indicate pixels that push the prediction away from that label.

The human evaluation of the explanations is presented in Figure 5. Overall, both approaches were rated positively, being perceived as clear, correct, useful, relevant, understandable, trustworthy, and relatively complete. A notable descriptive difference emerged with respect to simplicity: SPM was rated as simpler, whereas the ML approach was considered more complex and exhibited greater variability in participants' ratings. However, no statistically significant differences between the two approaches for any attribute in the semantic differential were observed ($p > 0.05$).

Experts rated the similarity between the SPM and ML approaches on a Likert scale, yielding a median score of 3.75 (median absolute deviation = 0.25). This indicates a relatively high level of perceived agreement between the two approaches. Regarding subjective preferences, 43.48% of the expert ratings favored the ML approach with SHAP explanations, followed by the SPM approach (34.78%) and the side-by-side presentation of both approaches (17.39%). Only 4.35% of ratings indicated no preference for any of the presented approaches.

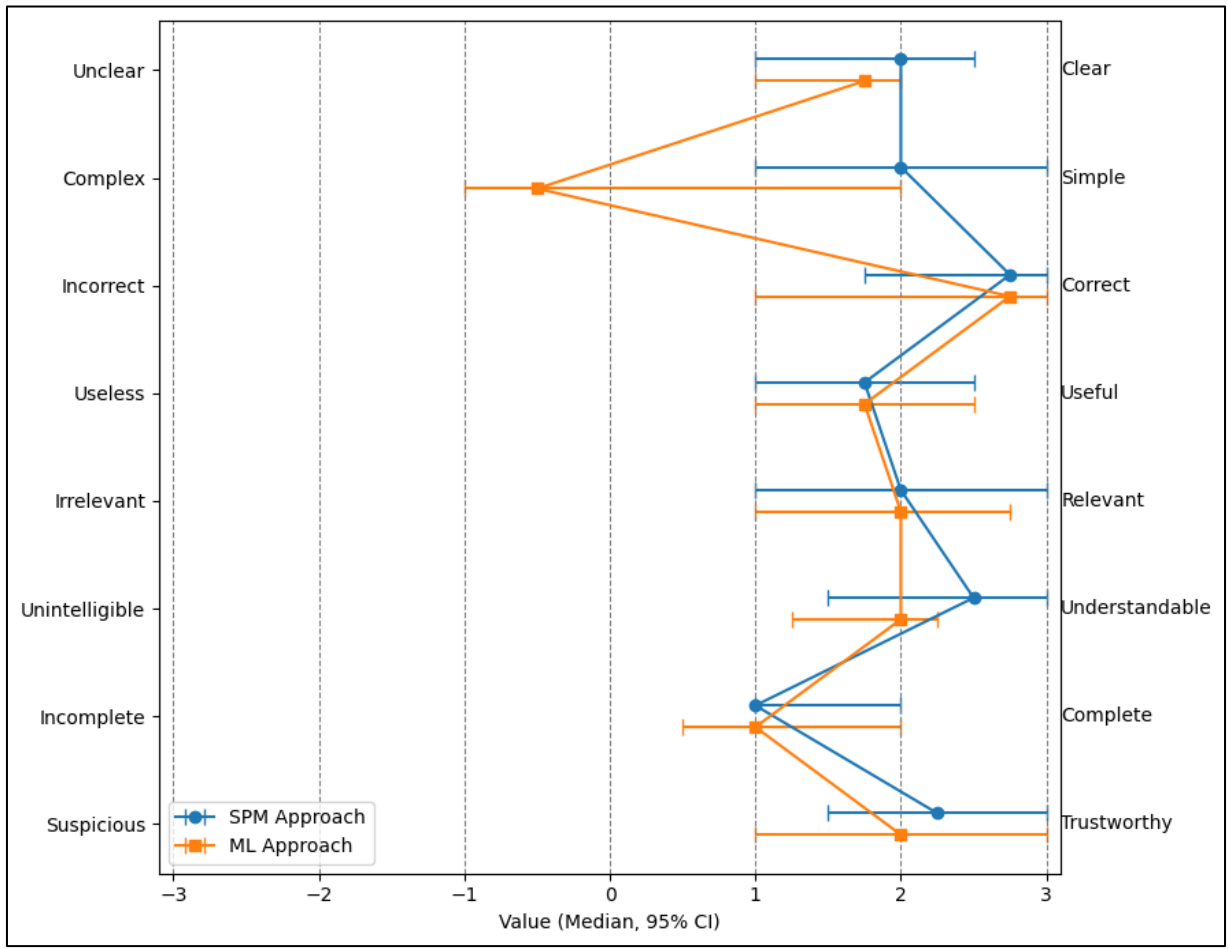

**Figure 5.** Results of the semantic differential survey, shown separately for the SPM (blue) and the ML (orange) approach. For clarity, positively connoted attributes are placed on the right side. Median values with 95% confidence intervals (CI; 5000 bootstrap samples) are shown.

## 4. Discussion

Both the SPM and ML approaches were able to detect outliers within plantar pressure data, but the ML approach outperformed the SPM approach in all evaluated metrics. Specifically, the ML model achieved an F1 score of 0.99 ± 0.00 and an MCC of 0.96 ± 0.01, compared to F1 = 0.93 ± 0.01 and MCC = 0.78 ± 0.02 for the SPM approach (research question 1). These results demonstrate the superior performance of the ML approach for outlier detection in this dataset. The observed differences in accuracy likely stem from the fundamentally different ways in which each approach interprets deviations. Within our dataset, clinically relevant but non-anomalous cases exist that exhibit substantial pixel-wise differences due to underlying pathological conditions or biomechanical adaptations. While these variations are clinically meaningful, they do not constitute technical errors or procedural inconsistencies. SPM is highly sensitive to localized, pixel-level variations and relies on precise normalization for rotation, scale, and anatomical landmarks to ensure that corresponding pixels accurately represent the same plantar region across subjects [17]. Although our proposed alignment procedure performed well visually, residual misalignments may have disproportionately affected SPM's classification performance.

This limitation is especially pronounced for pathological yet valid plantar pressure samples, where alignment with a normative reference plantar pressure distribution is challenging or even infeasible. For example, optimally registering a foot with Hallux Valgus to a normative template is nearly impossible, as either the hallux protrudes beyond the reference region or pressure voids appear where tissue is normally present. As a result, pathologic inliers were more often misclassified as outliers with the SPM approach due to their pixel-wise deviations from normative plantar pressure patterns. Regarding false negatives in the SPM approach, these largely arose from samples in outlier class 4 (*Incorrect Side Annotation*). This outcome may reflect the fact that lateral feet are relatively similar, and lateral differences are often localized and subtle. As a result, even when the lateral side is incorrectly labeled, the SPM approach may still interpret the sample as conforming to the normative dataset for that side, leading to misclassification.

The lowest label accuracy for the ML approach was observed for the outlier class *General Acquisition Error*, with 5.4% of the samples predominantly being wrongly classified as valid. A possible explanation is the class's inherent heterogeneity—ranging from subjects wearing shoes to instances with only partial plantar pressure data. Though combining these diverse characteristics into a single class was intended to boost sample size, it may have inadvertently compromised accuracy. Similar problems have been documented in other domains, such as medical imaging, where hidden stratification—arising from unrecognized heterogeneity within a class—has significantly reduced model performance [50]. Consequently, as the sample size of this outlier category grows, subdividing it into more homogeneous subclasses may enable finer-grained classification and improve outlier detection performance.

Overall, the two approaches were rated similarly on most attributes of the semantic differential, being generally rated positively across the dimensions, including clarity, correctness, usefulness, relevance, understandability, trustworthiness, and perceived completeness (research question 2). While the aggregated ratings suggest that experts generally found the explanations aligned with domain knowledge, individual cases reveal occasional differences in ratings, showing that full agreement with expert logic was not achieved for every sample. Nonetheless, the generally high usefulness ratings imply that both approaches could support expert understanding of the classification process, potentially facilitating interpretation of how specific features or regions contribute to model decisions. Taken together, these findings provide preliminary evidence that both approaches generate explanations that are interpretable and relevant from an expert perspective, though further investigation is needed to confirm the extent and robustness of this alignment.

The observed difference in perceived complexity was descriptive rather than statistically significant: although ratings exhibited high variability, SPM explanations were, on average, considered simpler. This descriptive trend aligns with expectations, as SHAP provides fine-grained, pixel-level attributions that detail how individual features contribute to predictions, whereas the SPM approach highlights only significant clusters of pixels, offering a less detailed but more immediately interpretable representation. The substantial variability in ratings, particularly for the ML approach using SHAP, likely contributed to the absence of statistical significance, reflecting the subjective nature of participants' perceptions of complexity.

The integration of ML-based and statistically driven explanations has been proposed as a promising avenue in XAI research, particularly in domains such as biomechanics where interpretability and trust are critical [16,51–53]. Previous findings suggest that the optimal XAI approach must be adapted to the user's context [54]. Consequently, providing both statistical and ML explanations allows end users to choose the representation that best fits their background and task requirements. Interestingly, our findings suggest that experts did not primarily value the combined presentation of SPM and ML explanations. Instead, the ML approach with SHAP explanations received the highest preference, followed by the SPM approach, while the side-by-side presentation of both approaches was less frequently favored. This indicates that, although SPM and ML operate at different levels of abstraction—feature-level versus group-level—the added value of presenting both simultaneously may not be as compelling to domain experts. At the same time, experts rated the overall similarity between the two approaches as relatively high, suggesting that despite methodological differences, both approaches were perceived as largely consistent. This perceived alignment may explain why experts felt comfortable selecting a single preferred approach rather than relying on a dual-validation perspective.

Finally, the necessity of XAI in the current study's task requires careful consideration. Although the classification task in this study is relatively straightforward for human experts, it is time-consuming, making automation valuable. In such scenarios, XAI primarily supports compliance with regulatory frameworks, fosters trust in automated systems, and facilitates human-in-the-loop monitoring [55]. It can also highlight cases where models fail, enhancing the robustness and reliability of ML-assisted decision-making. In our current evaluation, we focused on instances where SPM and SHAP explanations agreed with ground-truth labels. A practical workflow might prioritize ML predictions for decision-making, given their higher accuracy, while using SPM outputs as supporting explanations when the two approaches align.

This study has several limitations, which also suggest promising directions for future research. While our ML approach proved highly effective, simpler methods—such as expert-defined rules—could be explored for identifying specific outlier categories (e.g., multiple feet or upside-down feet). However, heuristic approaches tend to be less robust when handling highly abnormal foot shapes, as they may not adequately account for complex or anomalous data patterns [56]. In the context of inference-statistics–based outlier detection, future work could investigate advanced biomechanical alignment techniques, such as deformable image registration [57]. These approaches may better accommodate anatomical variability in pathological feet, improving registration quality and enhancing the robustness of SPM-based outlier detection. Moving beyond pixel-wise comparisons, aggregating plantar pressure values over anatomically or functionally meaningful regions (e.g., heel, metatarsal heads) could also provide more clinically meaningful and functional metrics for comparison with normative datasets [58]. Our dataset included both real and artificially generated outlier samples. Although experts confirmed that the synthetic cases closely resembled realistic anomalies, some residual bias cannot be ruled out. Future research could leverage generative AI to produce even more realistic artificial outliers, building on recent work demonstrating its capability to generate accurate biomechanical data and thus enhancing the utility of ML predictions in biomechanics [59–62].

Our study centered on SHAP as the XAI tool. Future work could systematically compare multiple XAI methods, assessing both their technical interpretability and perceived usefulness from the perspective of human experts. Expert ratings in our semantic differential analysis may have been influenced by differences in explanatory depth. Since standardized instruments for evaluating XAI explanations are still underdeveloped, our study represents an initial exploratory effort. Future research could refine evaluation protocols by expanding the set of assessment attributes and applying factor analysis to capture latent dimensions of user perception.

# 5. Conclusion

This study demonstrates that the statistical SPM analysis and ML modeling are highly promising approaches for detecting and categorizing technical errors and procedural inconsistencies in plantar pressure data, thereby enabling automated and targeted guidance for addressing anomalies. The results underscore that advancing artificial intelligence in biomechanics requires not only evaluating predictive performance but also understanding how users perceive model explanations. By applying semantic differential analysis to assess user perceptions of SPM and ML explanations, this study provides a first step toward developing human-centered tools that evaluate interpretability and practical usefulness, highlighting the need for further research in this direction.

## Data availability

## Author contributions

CaD: Conceptualization, Data curation, Formal Analysis, Funding acquisition, Methodology, Software, Validation, Visualization, Writing – original draft, Writing – review and editing. JD: Conceptualization, Validation, Writing – original draft, Writing – review and editing. SS: Writing – original draft, Writing – review and editing. DP: Conceptualization, Investigation, Writing – review and editing. SB: Writing – original draft, Writing – review and editing. HE: Conceptualization, Investigation, Writing – review and editing. KH: Conceptualization, Software. BS: Conceptualization, Writing – review and editing. ChD: Conceptualization, Investigation, Writing – review and editing. MF: Funding acquisition, Project administration, Supervision, Writing – review and editing.

## Funding

The authors declare that financial support was received for the research and/or publication of this article. This research was supported by the Central Innovation Program for Small and Medium-Sized Enterprises (Zentrales Innovations Program Mittelstand, ZIM) of the German Federal Ministry for Economic Affairs and Climate Action under Grant numbers 16KN113027 and KK5209402NK4.

## Competing interests

Authors HE, KH, and ChD were employed by DIERS International GmbH. The remaining authors declare that the research was conducted in the absence of any commercial or financial relationships that could be construed as a potential conflict of interest.